\documentclass[conference]{IEEEtran}
\IEEEoverridecommandlockouts
\usepackage{cite}
\usepackage{amsmath,amssymb,amsfonts}
\usepackage[noend]{algorithmic}
\usepackage{graphicx}
\usepackage{textcomp}
\usepackage{xcolor}
\usepackage{booktabs}
\usepackage{multirow}
\usepackage{microtype}
\usepackage{hyperref}
\usepackage{cleveref}

\usepackage{algorithm}

\newcommand{\eMCTSu}{Elastic MCTS$_u$}
\newcommand{\MCTSu}{MCTS$_u$}

\def\BibTeX{{\rm B\kern-.05em{\sc i\kern-.025em b}\kern-.08em
    T\kern-.1667em\lower.7ex\hbox{E}\kern-.125emX}}
\begin{document}

\title{Elastic Monte Carlo Tree Search \\
with State Abstraction for Strategy Game Playing
}

\author{\IEEEauthorblockN{Linjie Xu, Jorge Hurtado-Grueso, Dominic Jeurissen and Diego Perez Liebana}
\IEEEauthorblockA{\textit{Department of EECS} \\
\textit{Queen Mary University of London}\\
London, UK \\
\{linjie.xu,diego.perez\}@qmul.ac.uk}
\and
\IEEEauthorblockN{Alexander Dockhorn}
\IEEEauthorblockA{\textit{Faculty of EECS} \\
\textit{Leibniz University Hannover}\\
Hannover, Germany \\
dockhorn@tnt.uni-hannover.de} 
}

\maketitle

\begin{abstract}
Strategy video games challenge AI agents with their combinatorial search space caused by complex game elements. State abstraction is a popular technique that reduces the state space complexity. 
However, current state abstraction methods for games depend on domain knowledge, making their application to new games expensive. 
State abstraction methods that require no domain knowledge are studied extensively in the planning domain. However, no evidence shows they scale well with the complexity of strategy games. In this paper, we propose Elastic MCTS, an algorithm that uses state abstraction to play strategy games. 
In Elastic MCTS, the nodes of the tree are clustered dynamically, first grouped together progressively by state abstraction, and then separated when an iteration threshold is reached. The elastic changes benefit from efficient searching brought by state abstraction but avoid the negative influence of using state abstraction for the whole search. To evaluate our method, we make use of the general strategy games platform Stratega to generate scenarios of varying complexity. Results show that Elastic MCTS outperforms MCTS baselines with a large margin, while reducing the tree size by a factor of $10$. Code can be found at \url{https://github.com/egg-west/Stratega}
\end{abstract}


\section{Introduction}

The design of artificial intelligence (AI) for strategy game playing is a challenging problem. Human players can handle strategy games without much training when they possess knowledge about the game rules, opponent behavior, etc. AI solutions have different ways to gain knowledge about the game. Given domain knowledge, heuristics or scripts can be created. Heuristics can be used to guide the search of search-based AI agents~\cite{churchill2011build}. In combination with scripts, agents are able to deal with more complex strategy games~\cite{churchill2013portfolio}. 
However, those scripts are rigid and limited by the programmer's ability to capture the complexity of decision-making in strategy games (as most rule-based systems are). 
These scripts, once implemented, are static and they cannot adapt to new features introduced in the game, which is something typical during the game development cycle. In this case, these scripts need to be repeatedly updated to introduce new design updates.

An approach to get knowledge about the game is using a forward model. A forward model is a simulator that returns the next state given a state and an action. By simulating future states with the forward model, agents can sample different playing trajectories and store this knowledge in a data structure such as a tree~\cite{chaslot2008monte}. 
However, the size of the search space in strategy games increases combinatorially with the number of units under the control of the player, making existing search-based methods unable to find good solutions in a reduced time frame. Improving sampling efficiency is one of the main challenges for search-based methods in most domains (and, in particular in strategy games).

Game state and action abstraction~\cite{konidaris2019necessity} are efficient techniques that reduce the search space for search-based methods. Current works on state abstraction~\cite{uriarte2014game, uriarte2014high} and action abstraction~\cite{churchill2013portfolio, barriga2015puppet, barriga2017combining} have shown their improved performance over non-abstracting algorithms in playing StarCraft. Although these methods gain performance by utilizing abstraction techniques to reduce the search space, domain knowledge is indispensable for them. In the planning domain, methods that require no domain knowledge for abstraction are studied extensively~\cite{jiang2014improving, hostetler2014state, anand2015asap, hostetler2015progressive, anand2016oga, hostetler2017sample, sokota2021monte}. However, those have been limited to applications of lesser complexity, e.g. Othello~\cite{jiang2014improving}. In our study, we aim to test if similar concepts can be scaled up to strategy games.

In this paper, we propose an algorithm that employs state abstraction by approximate homomorphism~\cite{ravindran2004approximate} for Markov Decision Process (MDP). The generated state abstraction is used to merge tree nodes in Monte Carlo Tree Search (MCTS), which reduces the size of the tree.
This paper shows the challenges derived from implementing this approach and the solutions proposed to address them. One of these challenges is that, in order to obtain a good approximate homomorphism, a high number of samples is required. In fact, this number increases with large state and action spaces, which is problematic for strategy games where the state space and action space increases exponentially with the number of units and other elements of the game. 

We alleviate this problem in two ways. First, we implement a modification in MCTS (which we call \MCTSu) so that, for each node, the algorithm only considers the actions available for one unit (rather than for all of them). 
Secondly, we introduce an iteration threshold $\alpha_{ABS}\in \mathbb{N}$ that indicates a stopping iteration for the use of abstractions.
When the MCTS iteration number $N_{mcts}\in \mathbb{N}$ reaches $\alpha_{ABS}$, the state abstraction is abandoned and the tree is ``expanded'' again (abstract nodes are eliminated) to continue the search as in normal MCTS. Given the fact that the size of the tree changes during search, we call our algorithm Elastic MCTS.

Our contributions can be summarized as follows:
\begin{itemize}
    \item \textbf{Automatic state abstraction for strategy games with no domain knowledge:} Our method applies a state abstraction that requires no domain knowledge for complex environments such as strategy games, in contrast to  existing methods which require domain knowledge. While this work focuses on strategy games, the method proposed in this paper may be applicable to other genres, as it does not require game-specific knowledge. We assume the multi-unit setting in this paper while our method can be used for both single-unit and multi-unit settings.
    \item \textbf{An analysis on the effects of state abstraction on the recommendation policy:} Previous works keep the generated abstraction within the tree of MCTS during all iterations. This approach is based on the assumption that the policy resulting from the abstraction is better than the policy from the original state space, neglecting the risk of using a bad-quality state abstraction. Our algorithm sets up an iteration threshold for using the abstractions, which we tune to analyze the impact of turning back to the original tree at different times during the search. 
\end{itemize}

The rest of the paper is structured as follows: In Section~\ref{sec:background}, background knowledge about MCTS, state abstractions and approximate MDP homomorphism is introduced. The Stratega framework is introduced in Section~\ref{sec:stratega}. Section~\ref{sec:related_work} summarizes related works on state abstraction for strategy games and abstractions used in the context of planning. Section~\ref{sec:methods} describes Elastic MCTS, and in  Section~\ref{sec:experiments} we evaluate Elastic MCTS  empirically and compare it with other algorithms. Section~\ref{sec:conclusion_and_discussion} concludes our work and gives ideas for  future directions.

\section{Background}
\label{sec:background}

\subsection{Monte Carlo Tree Search (MCTS)} 

MCTS~\cite{chaslot2008monte} generates a search tree to estimate the state-action values of the current state. In this tree, each node represents a state and each branch represents an action, with the current state located at the root node. Each node stores the cumulative reward $X$ and the visit count $N$. Each MCTS iterations consists of 4 phases: selection, expansion, simulation, and backpropagation. 

During the selection phase, a tree policy is used to traverse the tree from the root until we reached a node on which we have not yet expanded all possible child nodes.
A popular choice for the tree policy is Upper Confidence Bounds (UCB) applied to Trees (UCT)~\cite{kocsis2006bandit}. For a node representing state $s$, connected with its parent who has an edge representing action $a$, its UCB value is:
\begin{align}
    UCB1(s,a) = \frac{X(s,a)}{N(s,a)} + C \sqrt{\frac{\ln{N_{\text{parent}}}}{N(s,a)}},
\end{align}
where the $X(s,a)$ is the cumulative reward, $N(s,a)$ is the visit count of this node, and $N_{parent}$ is the visit count of the parent node. A constant $C \in \mathbb{R}$ controls the trade-off between exploration (selecting nodes of low visit count) and exploitation (selecting known high-value nodes). The tree policy selects action with the highest UCB1 value, descending the tree until a non-fully-expanded node is reached. 

In the expansion phase, a child node is generated that represents the next state retrieved from applying a previously unexplored action. 
After expansion, it enters the simulation phase, where a roll-out policy will be used to continuously sample actions for a fixed amount of turns or until the end of the game. A classic roll-out policy is the random policy that chooses available actions uniformly. This final state is then evaluated by a state evaluation function and a score $R$ is backpropagated along the path taken during the selection and expansion phases. This value $R$ is normally the game outcome in terminal states, and a value returned by an heuristic function for non-terminal states.


\subsection{State Abstraction and approximate MDP Homomorphism}
\label{subsection:homomorphism}
A Markov Decision Process (MDP) is defined as \mbox{$<\mathbb{S}, \mathbb{A}, R, T, \gamma>$}, with state space $\mathbb{S}$, action space $\mathbb{A}$, reward function $R: \mathbb{S} \times \mathbb{A} \mapsto \mathbb{R}$, transition function $T: \mathbb{S} \times \mathbb{A} \times \mathbb{S} \mapsto \mathbb{R}$, and discount factor $\gamma \in \mathbb{R}, 0< \gamma < 1$ for discounting future rewards. 
State Abstraction for MDPs can be formalized as a mapping $\phi(s) = s_\phi, s\in \mathbb{S}, s_\phi \in \mathbb{S}_\phi$. The $\mathbb{S}_\phi$ is an abstract state space. Usually, we wish the size $|\mathbb{S}_\phi| < |\mathbb{S}|$ to reach a better sample efficiency or a shorter searching time. The approximate MDP homomorphism~\cite{ravindran2004approximate} defines the similarity between two states $s_1, s_2$ by defining two approximation errors:
\begin{align}
    \epsilon_R(s_1,s_2) = \max_a |R(s_1, a) - R(s_2, a)| \leq  \eta_R \label{eqn:jiang_eq}\\
    \epsilon_T(s_1,s_2) = \sum_{s'} |T(s'| s_1, a) - T(s'|s_2, a)| \leq \eta_T \label{eqn:jiang_eq2}
\end{align}
$\epsilon_R$ is the reward approximation error, while $\epsilon_T$ is the transition approximation error. $\eta_R$ and $\eta_T$ are the respective approximation thresholds. Two states are consider \textit{similar} if they hold that $\epsilon_R(s_1,s_2) \leq \eta_R$ and $\epsilon_T(s_1,s_2) \leq \eta_T$. In MCTS, \textit{similar} states from the same depth are considered candidates to construct a local approximate homomorphism. For each depth of the tree, two or more similar original (or \textit{ground}) states can be grouped into the same \textit{abstract state} or node. The reward and visiting count of the abstract node are $\hat{X} = \frac{\sum X_i}{m}$ and $\hat{N} = \frac{\sum N_i}{m}$, where $m$ is number of original nodes in this abstract node. When a new ground node is added to an abstract node, the statistics of this abstract node will be updated accordingly.

\section{Stratega}
\label{sec:stratega}

\begin{figure}
    \centering
    \includegraphics[width=0.48\textwidth]{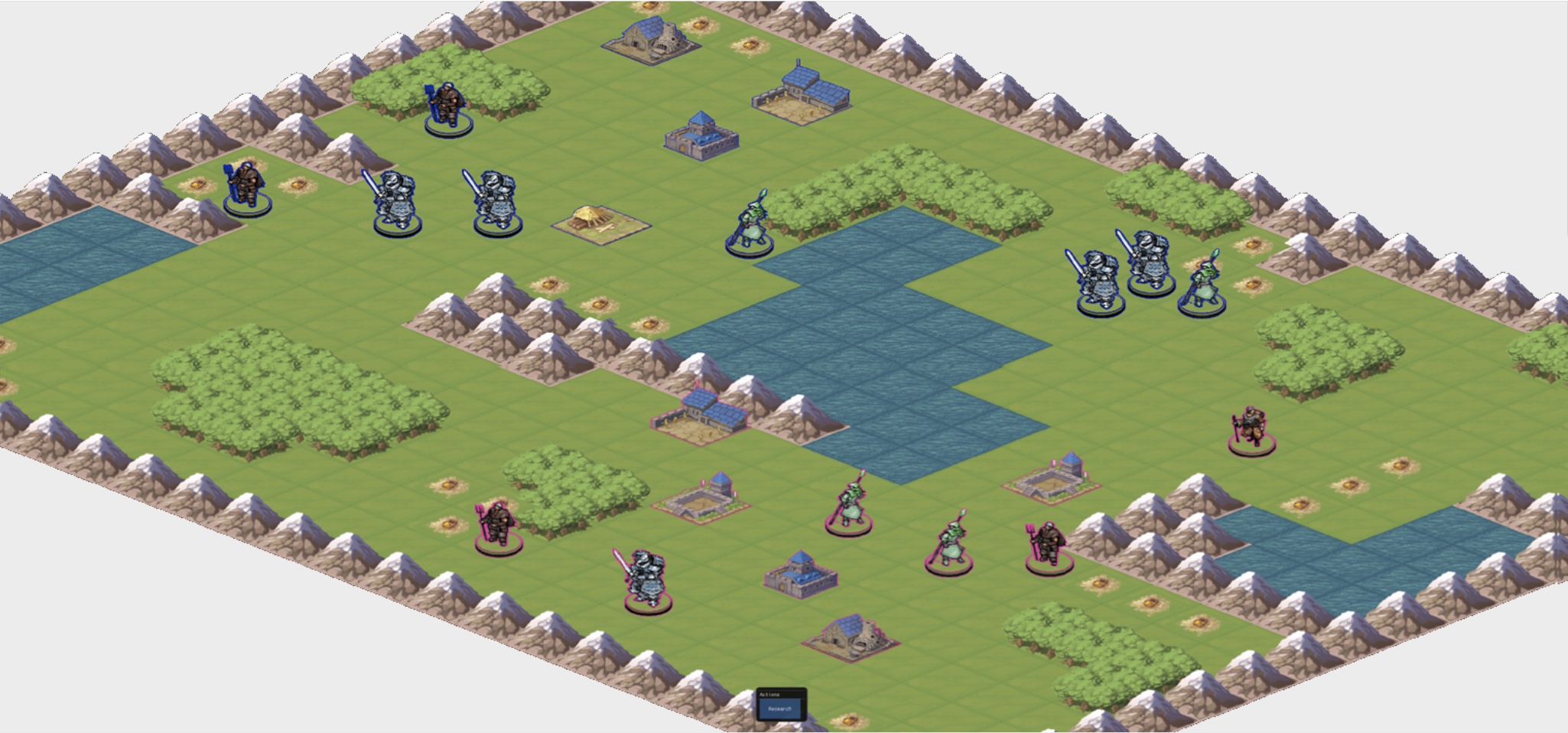}
    \caption{A classic scene of Stratega framework.}
    \label{fig:stratega}
\end{figure}

The Stratega~\cite{dockhorn2020stratega} framework (see screenshot in Figure~\ref{fig:stratega}) is developed for studying AI agents in general strategy game playing. Stratega uses an isometric view for the battlefield, where there are different tiles for game elements: landforms, buildings, resources, and army units. It allows developers to create their own turn-based and real-time strategy games through the YAML markup language, by setting up game elements and their parameters.
One of the important features of Stratega is that it provides a forward model that can be used by statistical forward planning methods, such as MCTS or Rolling Horizon Evolution~\cite{dockhorn2021portfolio}.

We use the game \textit{Kill The King} from Stratega for the evaluation of the search methods presented in this paper. \textit{Kill The King} is a turn-based, two-player strategy game where each player commands their units to defeat the opponent's king for a win. We limited the maximum number of turns to $100$. If after $100$ turns and no king dies, the game ends and returns a draw. In this game, we designed 4 different unit types: \textit{King, Warrior, Archer, Healer}. 
\textit{King, Warrior} and \textit{Archer} share the same action types: [\textit{Move, Attack, Do-nothing}]. The action types for \textit{Healer} are: [\textit{Move, Heal, Do-nothing}].
The action space of a unit depends on their action types and unit attributes. 
For example, the \textit{Move Range} of King is set to $2$, resulting in $12$ surrounding tiles for its \text{Move} action. 
Its \textit{Attack Range} is $2$, enabling it to choose any opponent unit to attack within 2 tiles. 
In this case, the maximum size of its action space is $(12+1) \times (12+1) = 169$, as units can first move ($+1$ for not moving) and then attack ($+1$ for not attacking). 
A player, at the beginning of their turn, can act with $n$ units, providing a combinatorial space bounded at $169^n$ for a turn.
In this paper, we experiment with a number of units between $4$ and $11$, which constitutes an action space bounded between $10^5$ and $10^{14}$ actions. The game is also complex from a strategic point of view: different unit types have distinct attribute values, which makes their behavior different and forces units to use different strategies. The attribute set for this game consists of: \textit{Health Points, Move Range, Attack Range, Attack Damage, Heal Strength}. The maps for Stratega are grid-like~\ref{fig:stratega}, where tiles such as mountains or water block the way, requiring an aspect of coordination between units for effective movement. 

\textit{Kill The King} is chosen to evaluate our proposed algorithm because of the following characteristics. Firstly, although it is not as complex as battles in Starcraft, these two games share challenges such as combinatorial state and action space, which is common in most strategy games. Second, the search space can be controlled by the number of units. Therefore, by increasing the number of units, we can test how the performance of our method scales with the increasing complexity of the search space and the size of the tree. Furthermore, varieties of the game can be created by changing the unit composition. These varieties are used to evaluate the methods by searching strategies for different environments.

\section{Related Work}
\label{sec:related_work}
State abstraction is a popular technique that shows its application in strategy games in the following works. \cite{chung2005monte} proposed a Monte Carlo planning algorithm to play \textit{Capture The Flag} game. 
To reduce the planning complexity, a handcrafted game state abstraction that divides the game map into tiles is used.
\cite{synnaeve2012bayesian} proposed a method that presents the map of StarCraft with regions connected by checkpoints, largely simplifying the state space. \cite{uriarte2014game} combines the state abstraction in~\cite{synnaeve2012bayesian} and action abstraction in playing Starcraft combats. 
Combat-irrelevant units such as workers and buildings are removed from high-level game state representation. With the search space reduced by their abstraction, their search-based method shows a performance close to a script-based agent. 
\cite{dockhorn2021game} proposes to encode the game state with vectors that contain information about entities. This representation enables eliminating superfluous information and grouping nodes with the same vector.

While the works mentioned above show that state abstraction is a powerful tool in complex searching space, \cite{uriarte2014high} investigates the effect of different state abstractions. In their work, 4 different state abstractions are created and they show different performance with MCTS in StarCraft. Among these works, different kinds of state abstraction methods are proposed. However, most of them require domain knowledge to construct the state abstraction. \cite{dockhorn2021game} utilizes a parameter optimizer to pick up entity information used for state abstraction, avoiding the use of human knowledge. However, this method shows no clear performance improvement in their evaluation. 

Most existing state abstraction applications in strategy games depend on domain knowledge. In the planning domain, state abstraction that requires no domain knowledge is more common. 
\cite{jiang2014improving} proposed approximate MDP homomorphism to construct state abstraction for MCTS. According to the state abstraction, states in each depth of MCTS that have a similar transition function and reward function are grouped. The similarity definition for this approach is shown in Eqs.~\ref{eqn:jiang_eq} and \ref{eqn:jiang_eq2}. The approximate homomorphism is generated from the samples collected by MCTS in a batch manner. Although their approach is shown to improve the performance in the planning domain, it requires extra action abstraction and pruning when applied to a board game such as Othello. 
\cite{anand2015asap} applies the approximate homomorphism to state-action abstraction, where the state-action pairs are grouped in MCTS.

\begin{figure*}
    \centering
    \includegraphics[width=0.85\textwidth,trim=0 1cm 0 1cm, clip]{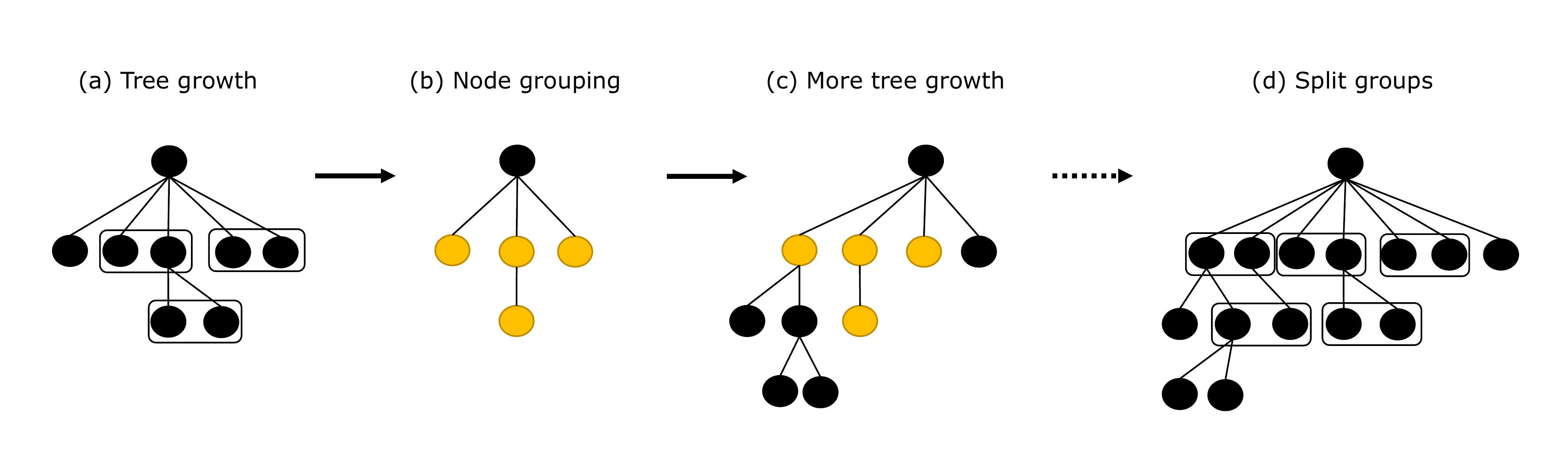}
    \caption{Overview of dynamic changes of tree nodes in Elastic MCTS. Ground nodes are \textit{black} while abstract nodes are \textit{yellow}. At first, the tree grows as in normal MCTS (a). After a number of $B$ iterations, ground nodes are grouped by using Approximate MDP Homomorphism (b). Search then continues adding more ground node (c), repeating state aggregation after every $B$ iterations. When the abstraction iteration threshold is reached, abstract nodes are split (d) and the search continues to explore the original game without using the state abstraction until the thinking budget expires.}
    \label{fig:elastic}
\end{figure*}

Progressive Abstraction Refinement for Sparse Sampling (PARSS)~\cite{hostetler2015progressive, hostetler2017sample} is a method that starts with a coarse state abstraction where all the states are clustered in the same group and refining it progressively. 
\cite{anand2016oga} proposed \textit{On-The-Go} abstraction that abandons the batch style of updating the abstraction~\cite{jiang2014improving, anand2015asap}. In their approach, the maintained abstraction is updated more frequently. In their work, a variable \textit{recent counts} for counting the number of visits is stored in each tree node. When it reaches a pre-defined threshold $\alpha_{otg}$, the abstraction for this state is updated and its \textit{recent counts} is reset. 
\cite{sokota2021monte} proposed Abstraction Refining that rejects adding a state that is similar to a known state in the tree. Their approach improves the performance of MCTS in stochastic environments. In conclusion, these methods are general and can be applied to different tasks. However, they are designed for planning domains and did not show if they can be applied to complex search spaces such as those from strategy games.

\section{Elastic MCTS}
\label{sec:methods}
We propose \textit{Elastic MCTS}, which combines approximate MDP homomorphism with MCTS, for strategy games. Our method is described in detail in Section~\ref{ssec:themethod} and depicted in Figure~\ref{fig:elastic}. Before this, we focus on two modifications that are needed to adapt the principles of MDP homomorphism to the large combinatorial search spaces present in this domain: unit ordering to reduce the action space (Section~\ref{ssec:prob1}) and introducing an iteration threshold to revert to the original (ground) state space.  


\subsection{Approximate Homomorphism in Strategy Games} \label{ssec:prob1}

Our method is based on state abstraction in the planning domain~\cite{jiang2014improving}, where a local approximate homomorphism (see \ref{subsection:homomorphism} for details) is constructed and constantly updated from trajectories sampled by MCTS. However, two issues occur when this method is applied to complex search spaces such as those from strategy games. 
The first issue is that the number of samples required to generate a good-quality approximate homomorphism depends on the sizes of the state and action spaces. In strategy games, where the state space is combinatorial and the action space changes according to different game states, the small number of samples collected within a limited decision budget results in a bad-quality state abstraction.

The second issue is that the reward and transition approximation errors $\epsilon_R$ and $\epsilon_T$ (Eqs.~\ref{eqn:jiang_eq} and~\ref{eqn:jiang_eq2}) are calculated executing all possible actions available in two states. For two states that have different action spaces, the original definitions of $\epsilon_R$ and $\epsilon_T$ conflict with the fact that most actions are likely not legal in \textit{both} states $s_1$ and $s_2$ at the same time. It is quite common in strategy games to observe different states with different sets of available actions. While it would be possible to resolve the approximation errors by only using the (small) set of common actions, the resulting values would not represent the true futures of both states in terms of rewards and transitions. Our initial tests (not included in this paper) showed that this indeed does not provide good abstractions for MCTS.

We alleviate these two issues by implementing a variant of MCTS called \MCTSu. In \MCTSu, each node corresponds to a state where a single unit can move. The node's edges are actions available to said unit only. Consequently, the action space for states candidate for merging is much smaller than the original combinatorial action space. Moreover, candidate states have a large proportion of common actions in their action space because they represent states where the same units act. 

A consideration for \MCTSu~is to decide the units' move order. In our implementation, the move order (by the sole purpose of the agent's search process) is set randomly at the beginning of the game and kept fixed during the whole game. While, theoretically, this ordering removes the guarantee of optimal convergence, empirically the advantage obtained by reducing the action space creates stronger players. This is shown in the results discussed in Section~\ref{sec:experiments} and is in line with previous results in the literature~\cite{7860427}. 

\subsection{Elastic State Grouping and Un-grouping}  \label{ssec:prob2}

We apply the constructed state abstraction to MCTS for grouping tree nodes. As~\cite{jiang2014improving}, the state abstraction in our method is also constructed in a batch manner: for every $B$ iterations of MCTS, the sampled trajectories are used to construct an approximate homomorphism, which aggregates similar nodes into groups according to the states they represent. Each MCTS node stores statistics including cumulative reward and visit count. For a node group, it stores these as the average of the statistics of all its ground nodes. The following MCTS iterations will be guided by these group statistics in two ways. First, the UCT value of one node for the selection is calculated based on the statistics of the group it belongs to. Second, when updating a node's statistic during back-propagation, their group nodes also update their statistics. 

Existing methods keep using the generated state abstraction until the search is finished. However, abstractions are known to introduce imperfections in the search space~\cite{anand2016oga}. With an erroneous abstraction, the policy obtained from abstraction performs worse than the policy derived from the ground search space. \MCTSu~reduces the size of action space and helps construct a better abstraction, but these imperfections remain. Another issue is that states in the same abstract node share their statistics while grouped,
forcing the action selection for the recommendation policy of MCTS to choose effectively at random among actions that lead to the same group node. 

We propose a novel approach to solve both problems mentioned above. We set up an iteration threshold for abstraction $\alpha_{ABS}$. After $\alpha_{ABS}$ iterations, MCTS assigns the node group statistics to each node in this group and abandons the state abstraction (breaking the node groups into ground tree nodes). The remaining MCTS iterations follow the normal MCTS algorithm. At this point, MCTS does not search in an imperfect space and nodes split from the same group might be now independently visited. The action-state values can become different and MCTS's recommendation policy can distinguish better among the available actions to suggest.

\subsection{Elastic MCTS} \label{ssec:themethod}

Algorithm~\ref{alg1} shows the pseudocode of Elastic MCTS, 
which runs MCTS iterations (line $3$) until the budget is exhausted (given in forward model calls, line $2$). 
The set of abstract nodes is initialized to set of the ground states (line $1$), and is updated after every batch of $B$ iterations (line $7$). When the number of iterations surpasses $\alpha_{ABS}$ (line $4$), we abandon state abstraction to return to the original ground search space (line $5$), 
by splitting all the nodes from state groups and assigning statistics of the state group to the ground nodes. This procedure is also depicted in Figure~\ref{fig:elastic}.

\begin{algorithm}[!t]
\caption{Elastic MCTS. 
$N_{fm} \in \mathbb{N}$: maximum forward model calls. $\alpha_{ABS} \in \mathbb{N}$: MCTS iteration threshold. $N_{mcts} \in \mathbb{N}$: current MCTS iteration number. $B \in \mathbb{N}$: batch size. $\phi(s) = \hat{s}$: state abstraction that maps an MCTS node representing $s$ to a node group $\hat{s}$. $\eta_R \in \mathbb{R}$ and $\eta_T \in \mathbb{R}$ are the reward function error threshold and transition error threshold, respectively.}
\label{alg1}
    \begin{algorithmic}[1]
    \REQUIRE $N_{fm} , \alpha_{ABS}, \eta_{R}, \eta_{T}$
    \STATE $\phi := s \rightarrow \hat{s}, \hat{s} = \{s\}$ \ \ \ \ \ \ \ \ \ \ \# Initialize the abstraction
    \WHILE{$USED\_FMCALL < N_{fm}$}

        \STATE $MCTSIteration(\phi)$
        
        \IF {$N_{mcts} > \alpha_{ABS}$}
            \STATE $\phi := s \rightarrow \hat{s}, \hat{s} = \{s\}$ \ \ \ \ \ \ \ \ \ \ \ \ \ \ \ \ \ \ \ \ \ \ \ \# Fig.\ref{fig:elastic} (d)
        \ELSIF{ $N_{mcts} \% B == 0$ }
            \STATE $\phi = ConstructAbstraction(\phi, \eta_{R}, \eta_{T})$\ \ \# Fig.\ref{fig:elastic} (b)
        \ENDIF
        \STATE $N_{mcts} = N_{mcts}+1$
    \ENDWHILE
    \end{algorithmic}
\end{algorithm}

Algorithm \ref{alg2} shows the MCTS iteration step. This only differs from normal MCTS in that it uses the statistics (cumulative reward and visit count) of the node group rather than the original tree nodes, for the selection (line $1$) and backpropagation (line $9$) steps. Note that new nodes added in the expansion phase are added as ground nodes to the tree, merging into states after $B$ iterations as shown in Algorithm~\ref{alg1}.

\begin{algorithm}[!t]
\caption{MCTSIteration($\phi$)}
\label{alg2}
\begin{algorithmic}[1]
    \WHILE{Select a child $K$ with maximum UCT value with $\phi$: $V_{uct} = X(\phi(s), a)/N(\phi(s), a) + C \sqrt{\ln {N_{parent}}/N(\phi(s),a)}$.}
        
        \IF {Node $K$ is not fully-expanded}
            \STATE Expand this node by generating a new child node $P$.
            \STATE Rollout for $P$ and obtain reward $R$ from state evaluation function.
            \STATE Break the while loop.
        \ELSIF{Node $K$ represents the end of game}
            \STATE Obtain reward $R$ from sate evaluation function.
            \STATE Break the while loop.
        \ENDIF
    \ENDWHILE
    \STATE Backpropagate $R$, updating $X(s,a)$ and $N(s,a)$ for node group $\phi(s)$ in selection path.
\end{algorithmic}
\end{algorithm}

Algorithm \ref{alg3} updates the state abstraction: from leaf nodes to the root (line $1$), for every ground tree node that is not part of an abstract node (line $2$), a similarity check is performed against all sibling abstract nodes. This similarity is determined by computing the values of two errors $\epsilon_R$ and $\epsilon_T$ from samples. 
The samples from MCTS are triplets that consist of a state, an action and a return: $<s, a, R>$. For computing the $\epsilon_R$ error between two states $s_1$ and $s_2$, we calculate $|R(s_1,a) - R(s_2, a)|$ for each action $a$ present in either action space of $s_1$ and $s_2$, with  
$R(s_i, a) = 0$ if $a$ is invalid for any $s_i$.
$\epsilon_R$ the takes the value of the maximum difference found (line $5$).
$\epsilon_T$ is also calculated for all actions (line $6$), with $|T(s'|s_1, a) - T(s'|s_2, a)| = 0$ only when the $s_1$ and $s_2$ both have action $a$ available and this action leads to the same next state $s'$. Because we only consider two state representing the same unit for state abstraction, their next state $s'_1$ and $s'_2$ are recognized the same when the unit-related attributes have the same values.

If an abstract node is found where $\epsilon_R \leq \eta_R $ and $ \epsilon_T \leq \eta_T$, we add the ground node to this node (lines $7-8$). If no group fulfills this condition, we create a new group including only the ground node in it (line $10$).

\begin{algorithm}[!t]
\caption{ConstructAbstraction($\phi, \eta_{R}, \eta_{T}$), $l$ is the tree depth and $L$ is the maximum depth of the current tree.}
\label{alg3}
\begin{algorithmic}[1]
    \FOR{$l=L$ to $1$}
        \FORALL{state $s_1$ in depth $l$ that is not grouped}
            \FORALL{abstract state $\hat{s}$ in $\phi$}
                \STATE $s_1$\_in\_$\hat{s}$ $=$ true
                \FORALL{state $s_2$ in $\hat{s}$}
                \STATE $\epsilon_R = \max_a |R(s_1, a) - R(s_2, a)| $
                \STATE $\epsilon_T = \sum_{s'} |T(s'|s_1, a) - T(s'|s_2, a)|$
                \IF { $\epsilon_R > \eta_R $ \textbf{or} $ \epsilon_T > \eta_T $}
                \STATE $s_1$\_in\_$\hat{s}$ $=$ false, \textbf{break}
                \ENDIF
                \ENDFOR
            \IF { $s_1$\_in\_$\hat{s}$ $==$ true}
            \STATE Add  $s_1$ in abstraction node
            \ELSE 
            \STATE Create a new abstract node
            \ENDIF
            \ENDFOR
        \ENDFOR
    \ENDFOR
    
\end{algorithmic}
\end{algorithm}

\section{Experiments}
\label{sec:experiments}

Four agents are used for the experiments in this paper:
\begin{enumerate}
    \item \textit{Combat Agent}: This agent is based on a rule-based agent~\cite{dockhorn2020stratega} built-in Stratega. Its strategy is to concentrate attacks to a single isolated enemy unit and assign healers to heal the strongest ally units. The isolation score for a unit depends on the number of nearby ally and enemy units. 
    Once a target is chosen, the agent searches available action for each agent to i) get close to the target, ii-a) attack the target or ii-b) heal the target. 
    \item \textit{MCTS}: the default MCTS algorithm with no abstraction nor unit ordering.
    \item \textit{\MCTSu}: MCTS with unit ordering, no state abstraction.
    \item \textit{\eMCTSu}: Elastic MCTS with fixed unit ordering.
\end{enumerate}

The same \textit{state evaluation function} is used by \textit{MCTS}, \MCTSu~and \eMCTSu. This function gives an utility value $0.0 \leq R \leq 1.0$ to a state after normalization. The utility values for win, loss and draw are $1, -1, 0$, respectively.
If the given state is not terminal, the value of the state is $R = 1-\frac{d h}{DH}$, where $d$ is the distance between player's units and opponent's king, $h$ the health point of opponent's king, and $D$ and $H$ are the maximum values of $d$ and $h$. In conclusion, our state evaluation function encourages the agent to get close to opponent's king and attack it. 

\subsection{Parameter Optimization for Agents with NTBEA}
\label{subsection:parameter}

All our agents are pitched against each other in $1 vs 1$ games. We use the optimiser N-Tuple Bandit Evolutionary Algorithm (NTBEA)~\cite{lucas2018n} to tune the parameters of each agent. NTBEA utilizes a combinatorial multi-armed bandit to navigate the parameter space, while building a landscape model for a noisy evaluation function.

For NTBEA, we set the exploration factor for the multi-armed bandit to $2$, we use $50$ neighbors, and a limit of $50$ iterations. When tuning the agent parameters, the fitness is set as the win rate of the agent playing against \textit{CombatAgent}. In these and all the following experiments, the computational budget for each action decision by all search agents is set to $30,000$ forward model calls. As an opponent model, all search agents simply pick valid actions uniformly at random. Additionally, for \eMCTSu, we empirically determined the error approximation thresholds for $\eta_R = 0.1$ and $\eta_T = 0.3$.

The following describes the parameter space explored by NTBEA: MCTS and \MCTSu~have the same parameter spaces: we evaluate the exploration factor $C \in \{0.1, 1, 10, 100\}$ and rollout length $L \in \{20, 40, 60, 80, 100\}$.
\eMCTSu~adds to these two parameters ($C$ and $L$) the batch size $B \in \{20, 40, 60\}$ and the iteration threshold to stop using abstractions $\alpha_{ABS} \in \{4\times B, 8\times B, 12\times B, 16 \times B\}$. 
The parameters found are $\{C=0.1, L=20\}, \{C=10, L=100\}$ and $\{C=0.1, L=40, B=20, \alpha_{ABS}=12\}$ for MCTS, \MCTSu~and \eMCTSu.


\subsection{Algorithmic Performance}
\label{subsection:general}

Using the parameter values as tuned by NTBEA, agents are evaluated by playing against each other in different variants of $Kill The King$. We designed two groups of experiments to evaluate the performance of the proposed method. The first group evaluates the presented agents across scenarios with different amount of units, to observe how this performance changes as the action space becomes larger. The second group investigates the performance in maps with different layouts. Win Rates for both players are reported  (draw rates can be inferred). Each win rate is shown with its standard deviation, obtained by running each pairing $5$ times with $5$ different seeds on the same setting (game map, starting unit positions, etc.).

In the experiment for different unit numbers, we have 3 army compositions: \textit{(1 King, 1 Warrior, 1 Archer, 1 Healer), (1 King, 2 Warriors, 2 Archers, 2 Healers) and (1 King, 3 Warriors, 3 Archers, 3 Healers)}. For each army composition, $50$ game levels based on the map ``lak110d''~\cite{sturtevant2012benchmarks}
are generated by randomly choosing initial positions for each unit. In total, there are $500$ games played for each army composition ($5$ seeds, $50$ game levels, $\times 2$ from switching starting sides). 

Tables~\ref{Tab:general_k1w1a1h}, \ref{Tab:general_k2w2a2h} and \ref{Tab:general_k3w3a3h} show the experimental results. As can be seen in Table~\ref{Tab:general_k1w1a1h}, all three MCTS, \MCTSu~and \eMCTSu~ clearly outperform the \textit{Combat Agent}. When playing against MCTS, \MCTSu~and \eMCTSu~both outperform MCTS by a large margin. The win rate of \MCTSu~is slightly higher than the win rate of \eMCTSu, but the difference is at a scale of the standard deviation. In the results of games played between \eMCTSu~and \MCTSu, we can see a higher winning rate for \eMCTSu, with a difference larger than $10.0\%$. This consistently shows the improvement obtained by introducing state abstraction to the algorithm, with \eMCTSu~showing good performance against all other agents. 

When the search complexity increases from $4$ units to $7$ units (Table~\ref{Tab:general_k2w2a2h}) and $10$ units (Table~\ref{Tab:general_k3w3a3h}), the results mentioned above remain consistent. It is worth noting that the Elastic \MCTSu~performs better when the number of units goes up (see results when playing against \textit{Combat Agent} and \MCTSu), which indicates that our method for state abstraction is able to scale appropriately when using more units. 

\begin{table}[]
\caption{Win rates with standard deviation for games with 1 King, 1 warrior, 1 archer and 1 healer.}
\centering
\begin{tabular}{rr|cc}
\toprule
Agent 1 & Agent 2 & Agent 1 & Agent 2 \\ \midrule
MCTS & Combat Agent                  &   $54.6\pm4.7\%$    &   $45.4\pm4.7\% $                 \\ 
\MCTSu & Combat Agent   &   $67.2\pm2.9\%$    &    $32.6\pm2.9\%$                 \\ 
\eMCTSu& Combat Agent   &       $71.3\pm3.1\%$         &    $28.7\pm3.1\%$                 \\ 
\MCTSu& MCTS           &       $66.0\pm5.7\%$     &        $25.8\pm5.0\%$            \\ 
\eMCTSu& MCTS     &     $65.8\pm4.0\%$        &     $30.4\pm4.4\%$          \\ 
\eMCTSu& \MCTSu     &   $51.6\pm6.4\%$    &    $40.8\pm8.6\%$             \\ \bottomrule
\end{tabular}
\label{Tab:general_k1w1a1h}
\end{table}

\begin{table}[]
\caption{Win rates with standard deviation for games with 1 King, 2 warriors, 2 archers and 2 healers}
\centering
\begin{tabular}{rr|cc}
\toprule
Agent 1 & Agent 2 & Agent 1 & Agent 2 \\ \midrule
MCTS& Combat Agent                  &   $55.8\pm3.2\%$    &   $44.2\pm3.2\%$                  \\ 
\MCTSu& Combat Agent   &   $74.4\pm2.4\%$    &    $25.6\pm2.4\%$                 \\ 
\eMCTSu& Combat Agent   &       $77.4\pm2.1$\%         &    $22.4\pm1.7\%$                 \\ 
\MCTSu& MCTS           &       $72.0\pm2.8\%$     &        $21.8\pm2.4\%$            \\ 
\eMCTSu& MCTS     &     $69.0\pm4.5\%$        &     $27.0\pm3.0\%$          \\ 
\eMCTSu& \MCTSu     &   $54.4\pm3.2$\%    &    $37.0\pm3.4\%$             \\ \bottomrule
\end{tabular}
\label{Tab:general_k2w2a2h}
\end{table}

\begin{table}[]
\caption{Win rates with standard deviation for games with 1 King, 3 warriors, 3 archers and 3 healers}
\centering
\begin{tabular}{rr|cc}
\toprule
Agent 1 & Agent 2 & Agent 1 & Agent 2 \\ \midrule
MCTS& Combat Agent                  &   $59.6\pm2.6\%$    &   $40.4\pm2.6\% $                 \\ 
\MCTSu& Combat Agent   &   $84.8\pm2.5\%$    &    $15.2\pm2.5\%$                 \\ 
\eMCTSu& Combat Agent   &       $82.6\pm2.3$\%         &    $17.4\pm2.3\%$                 \\ 
\MCTSu& MCTS           &       $77.6\pm3.1\%$     &        $15.2\pm3.1\%$            \\ 
\eMCTSu& MCTS     &     $75.8\pm2.5\%$        &     $20.8\pm2.3\% $         \\ 
\eMCTSu& \MCTSu     &   $49.1\pm1.9\%$    &    $38.8\pm2.7\%$             \\ \bottomrule
\end{tabular}
\label{Tab:general_k3w3a3h}
\end{table}

To evaluate the performance of the proposed method in different maps, we choose 27 maps from~\cite{sturtevant2012benchmarks}. 
5 army compositions from 4 to 11 units are tested in each map (see Table~\ref{Tab:layout}). We evaluate the performance of \eMCTSu~playing against \MCTSu. Table~\ref{Tab:layout} shows the average win rate with its corresponding standard deviation for both agents in each army composition. 
In all games, \eMCTSu~outperforms \MCTSu~by large margins (between $19.6\%$ and $34.1\%$). 
These results show that the proposed method improves the performance of the algorithm and its improvements are consistent in different army compositions and game maps.

\begin{table}[t]
\caption{Win rates with standard deviation for games where \eMCTSu~played against \MCTSu~in 27 different layouts. The army composition column indicates the number of warriors (\textit{X}W), archers (\textit{X}A), healers (\textit{X}H) and 1 king (K).}
\centering
\begin{tabular}{rr|cc}
\toprule
Army Composition & \eMCTSu & \MCTSu \\ \midrule
K3H         &   $61.1\pm8.3\%$    &   $33.3\pm7.4\% $                 \\ 
K3W3A3H   &     $51.1\pm6.8\%$  &       $30.0\pm4.1\%$              \\ 
K10A   &       $59.3\pm5.5\%$         &    $25.2\pm3.8\%$                \\ 
K10W           &   $54.8\pm4.9\%$      &        $35.2\pm6.1\%$            \\ 
K5W5A     &     $53.0\pm4.5\%$       &     $30.7\pm5.7\%$        \\ 
\bottomrule
\end{tabular}
\label{Tab:layout}
\end{table}

\subsection{Influence of Abstraction Threshold}

To investigate the influence of different iteration thresholds, we pitch \eMCTSu~against \MCTSu, with the map settings and army compositions used for the experiments whose results are shown in Tables~\ref{Tab:general_k1w1a1h}, \ref{Tab:general_k2w2a2h} and \ref{Tab:general_k3w3a3h}. 
The values used for the agents' parameters are those obtained by NTBEA, as mentioned in Section~\ref{subsection:parameter}, with the exception of the iteration threshold $\alpha_{ABS}$. The values of this parameter are explored in this experiment to observe the effect of using state abstraction during different proportions of MCTS iterations. We try with proportions $\alpha_{ABS} \in \{0\%, 25\%, 50\%, 75\%, 100\%\}$. 
When it is set to $0\%$, the algorithm behaves like (normal) MCTS. When it is $100\%$, the state abstraction is used during all MCTS iterations and the recommendation policy picks an action based on statistics from the group nodes. 

As Table~\ref{Tab:tab4} shows, the performance of the \eMCTSu~agent differs when the abstraction threshold changes. 
Note these two agents use different hyper-parameters (tuned by NTBEA) thus their win rates differ with proportion set to $0\%$. 
There seems to be a sweet spot on the value of this threshold: performance is best when using the state abstraction during the first $50\%$ and $75\%$ iterations of \eMCTSu, obtaining lower winning rates when closer to $0\%$ and $100\%$. This result is especially noticeable for a threshold value of $100\%$, where the win rate is the lowest one for all army compositions. In all these, \MCTSu~obtains a consistently higher winning rate against \eMCTSu. 
Our interpretation of this phenomenon is that nodes sharing the same statistics in abstracted nodes until the end of the decision process makes the recommendation policy of \eMCTSu~behave suboptimally, as there is not enough information to discern between the different available actions at the root node. However, abandoning the abstraction at an intermediate point during the search lands better results than \MCTSu~also in all army compositions, showing the usefulness of Elastic MCTS and these abstractions.

\begin{table}[]
\caption{
Win rates with standard deviation of \eMCTSu~vs \MCTSu.
The proportion column indicates at which \% of the search state abstraction is abandoned.
}
\centering
\begin{tabular}{cr|r|r}
\toprule
Army Composition              & Proportion & \eMCTSu & \MCTSu \\ \midrule
\multirow{5}{*}{KWAH}    & 0\%        & \multicolumn{1}{r|}{$47.4\pm2.4\%$} &   $46.0\pm3.8\%$     \\ 
                         & 25\%       &        $50.0\pm2.7\%$    &    $40.4\pm3.1\%$       \\ 
                         & 50\%       &        $51.6\pm4.6\%$    &  $39.4\pm3.8\%$     \\ 
                         & 75\%       & $\mathbf{51.8\pm4.4\%}$  &     $\mathbf{38.2\pm6.0\%}$     \\ 
                         & 100\%      &        $39.0\pm2.8\%$    &    $45.0\pm1.8\%$  \\ \midrule
\multirow{5}{*}{K2W2A2H} & 0\%        & \multicolumn{1}{r|}{$49.0\pm3.6\%$}   & $42.6\pm4.0\%$  \\ 
                         & 25\%       &         $49.8\pm3.9\%$    &    $42.6\pm4.8\%$ \\ 
                         & 50\%       &         $\mathbf{52.8\pm4.8\%}$  &  $\mathbf{37.6\pm4.1\%}$  \\ 
                         & 75\%       &        $52.8\pm3.6\%$  &  $39.2\pm6.4\%$ \\ 
                         & 100\%      &        $38.8\pm3.3\%$       &   $47.2\pm1.5\%$  \\ \midrule
\multirow{5}{*}{K3W3A3H} & 0\%        & 
\multicolumn{1}{r|}{$49.0\pm 3.8\%$}  &  $38.8\pm3.5\%$   \\ 
                         & 25\%       &       $47.8\pm 4.3\%$       &  $39.0 \pm 3.8\%$ \\ 
                         & 50\%       &        $\mathbf{55.3\pm3.9\%}$       &   $\mathbf{33.8 \pm 3.3\%}$  \\ 
                         & 75\%       &        $49.2 \pm 3.3\%$     &   $40.2\pm 3.1\%$        \\ 
                         & 100\%      &      $40.0\pm2.3\%$     &   $44.4\pm1.4\%$   \\ \bottomrule
\end{tabular}
\label{Tab:tab4}
\end{table}

\subsection{Compression Rate}

\begin{figure} 
    \centering
    \includegraphics[width=0.6\columnwidth,trim=0 0cm 0 0cm, clip]{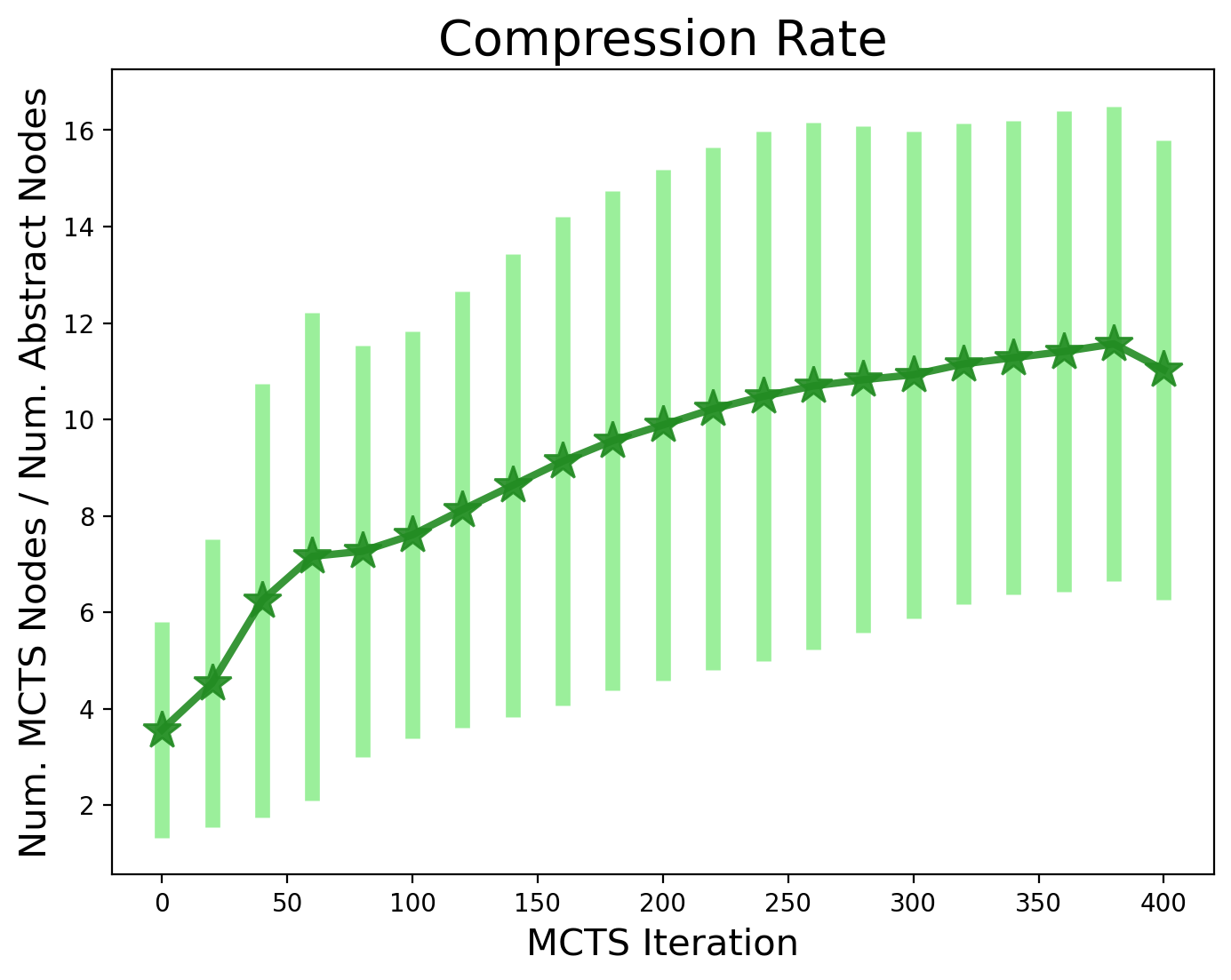}
    \caption{Compression Rates with standard deviation from 40 game plays.}
    \label{fig:compression}
\end{figure}

To show the difference between the ground and the abstracted state space, we define a compression rate as $N_{tree} / N_{abs\_tree}$, where $N_{tree}$ is the number of nodes generated by MCTS and $N_{abs\_tree}$ is the number of node groups generated by the abstraction. 
Note that with the same tree from MCTS, fewer groups mean more nodes are grouped together and the size of the tree is reduced more. We evaluate this compression rate in 20 instances of the map ``lak110d'' with the army composition \textit{(1 King, 1 Warrior, 1 Archer, 1 Healer)}. 
Figure~\ref{fig:compression} visualizes the achieved compression rate against MCTS iterations, showing a moderate increase of compression rate over time. 
In our previous experiments, the state abstraction was abandoned in the $240$th iteration ($\alpha_{ABS} \times B = 12 \times 20$), corresponding to a compression rate of $10$ states per group node. 

The state abstraction does not influence the computing time significantly. The average decision times are $667\pm 13$ and $685\pm 13$ ms for MCTS$_u$ and  Elastic MCTS$_u$, respectively.

\section{Conclusion and Future Work}
\label{sec:conclusion_and_discussion}

In this paper, we propose a variant of MCTS that generates clusters of nodes by using state abstraction. 
Our work is inspired by methods that use state abstraction in planning domains, but it is adapted to target complex search and action spaces, while still requiring no domain knowledge.
Our experiments show that the proposed method outperforms MCTS and a baseline rule-based agent in multiple variants of a turn-based strategy game, \textit{Kill the King}. 
We also observe a reduction of the tree size by a factor of $10$ and a considerable improvement in the agent's performance when using the state abstraction during a proportion of the search $<100\%$. 

The results shown in this paper have been obtained in a turn-based game, but we hypothesize that the gains in performance may also benefit other real-time or more complex turn-based strategy games. Due to the increased number of units controlled in games like Starcraft and the restricted time to return an action, we believe that the observed efficiency improvements may transfer well to RTS games. This is an immediate case for future work, where we'll investigate the applicability and scalability of Elastic MCTS to more complex games.

Besides aiming to improve the performance of the game playing agent, other aspects can also be investigated. For example, it is interesting to observe if different gameplay styles can be obtained when using these abstractions or if, on the contrary, the simplifications made in the tree prevent us from that goal. Recent works on quality-diversity methods applied to strategy games could be explored in conjunction with Elastic MCTS~\cite{perez2021generating}. Additionally, one of the most widespread methods for strategy games and large action spaces is the use of portfolio methods~\cite{churchill2013portfolio}, and it will be interesting to see the potential synergies of state abstraction with these algorithms. Finally, it is also possible to try different mechanisms for state abstraction (e.g.~\cite{hostetler2015progressive}), which could be an alternative to Approximante MDP Homomorphism. 
In fact, considering the complexity of strategy games, it's possible that different abstractions can be applicable to different moments of the game, which lies an interesting line of research ahead to discover which abstraction methods can be applied to which game situations.  

\section*{Acknowledgments}
Work supported by UK EPSRC grant EP/T008962/1.

\bibliographystyle{unsrt}
\bibliography{elastic_cog}

\end{document}